\documentclass[journal]{IEEEtran}

\usepackage{graphicx}  %insert graph
\usepackage{epstopdf}  %insert eps file
\usepackage{multirow}
\usepackage{array}  %assign the width
\usepackage{color}
\usepackage{colortbl}
\usepackage{booktabs}
\usepackage{amsmath}
\usepackage{mathrsfs}  %use curlicue letters
\usepackage{amsfonts}  %use hollow body letters
\usepackage[switch]{lineno}
\usepackage{algorithm}
\usepackage{algpseudocode}
\usepackage{amsmath}
\usepackage{booktabs}
\usepackage{threeparttable}
\usepackage{url}

\begin{document}
	
\title{Revisiting Anchor Mechanisms for Temporal Action Localization}
\author{Le Yang, Houwen Peng, Dingwen Zhang~\IEEEmembership{Member,~IEEE} \\ Jianlong Fu~\IEEEmembership{Member,~IEEE}, Junwei Han,~\IEEEmembership{Senior Member,~IEEE}% <-this % stops a space
\thanks{This work was supported by the Key-Area Research and Development Program of Guangdong Province(2019B010110001), the Innovation Foundation for Doctor Dissertation of Northwestern Polytechnical University(CX201916), the Research Funds for Interdisciplinary subject NWPU, the China Postdoctoral Support Scheme for Innovative Talents under Grant BX20180236, the National Natural Science Foundation of China under Grants 61876140 and U1801265. (\emph{Corresponding author: Junwei Han.})}% <-this % stops a space
\thanks{L. Yang, D. Zhang and J. Han are with the School of Automation, Northwestern Polytechnical University. D. Zhang is also with the School of Mechano-Electronic Engineering, Xidian University. H. Peng and J. Fu are with Microsoft Research Asia. (e-mail: junweihan2010@gmail.com).}% <-this % stops a space
\thanks{Code available at: https://github.com/LeYangNwpu/A2Net.}% <-this % stops a space
}
\markboth{IEEE TRANSACTIONS ON IMAGE PROCESSING, VOL. -, NO. -, - 2020}%
{Shell \MakeLowercase{\textit{et al.}}: Bare Demo of IEEEtran.cls for IEEE Journals}

\maketitle

\begin{abstract}
	Most of the current action localization methods follow an anchor-based pipeline: depicting action instances by pre-defined anchors, learning to select the anchors closest to the ground truth, and predicting the confidence of anchors with refinements. Pre-defined anchors set prior about the location and duration for action instances, which facilitates the localization for common action instances but limits the flexibility for tackling action instances with drastic varieties, especially for extremely short or extremely long ones. To address this problem, this paper proposes a novel anchor-free action localization module that assists action localization by temporal points. Specifically, this module represents an action instance as a point with its distances to the starting boundary and ending boundary, alleviating the pre-defined anchor restrictions in terms of action localization and duration. The proposed anchor-free module is capable of predicting the action instances whose duration is either extremely short or extremely long. By combining the proposed anchor-free module with a conventional anchor-based module, we propose a novel action localization framework, called A2Net. The cooperation between anchor-free and anchor-based modules achieves superior performance to the state-of-the-art on THUMOS14 (45.5\% vs. 42.8\%). 	Furthermore, comprehensive experiments demonstrate the complementarity between the anchor-free and the anchor-based module, making A2Net simple but effective.
\end{abstract}


\IEEEpeerreviewmaketitle

%%%%%%%%% BODY TEXT
\section{Introduction}

\IEEEPARstart{T}{he} temporal action localization task aims to determine the start and end time of an action instance as well as the corresponding category label. It can detect informative fragments from the given untrimmed videos, facilitating applications such as smart surveillance, highlight extraction, video summary, etc. \cite{gaidon2013temporal, wei2019sequence}.

\begin{table}[tbp]
  \caption{The performance of anchor-free module and anchor-based module for actions with different durations on THUMOS14 dataset, measured by mAP under threshold $0.5$.}
  \label{tab_subdataset_performance}
  \centering
  \small
  \begin{threeparttable}
      \begin{tabular}{c|ccccc}
      	\toprule[2pt]
        Sub-dataset     &ES\tnote{*}	        &Short	        &Medium	        &Long      &EL\tnote{**} \\
        \midrule[1pt]
        anchor-free 	&\textbf{14.52}		&\textbf{28.08	}	&33.73		        &47.68	        &\textbf{34.36}\\
        anchor-based 	&5.95		        &24.25		        &\textbf{51.60	}	&\textbf{63.86}	&31.84\\
        \bottomrule[2pt]
      \end{tabular}
      \begin{tablenotes}
        \footnotesize
        \item[*]  Extremely Short
        \item[**] Extremely Long
      \end{tablenotes}
  \end{threeparttable}
\end{table}

\begin{figure}[thbp]
  \graphicspath{{figures/}}
  \centering
  \includegraphics[width=0.9\linewidth]{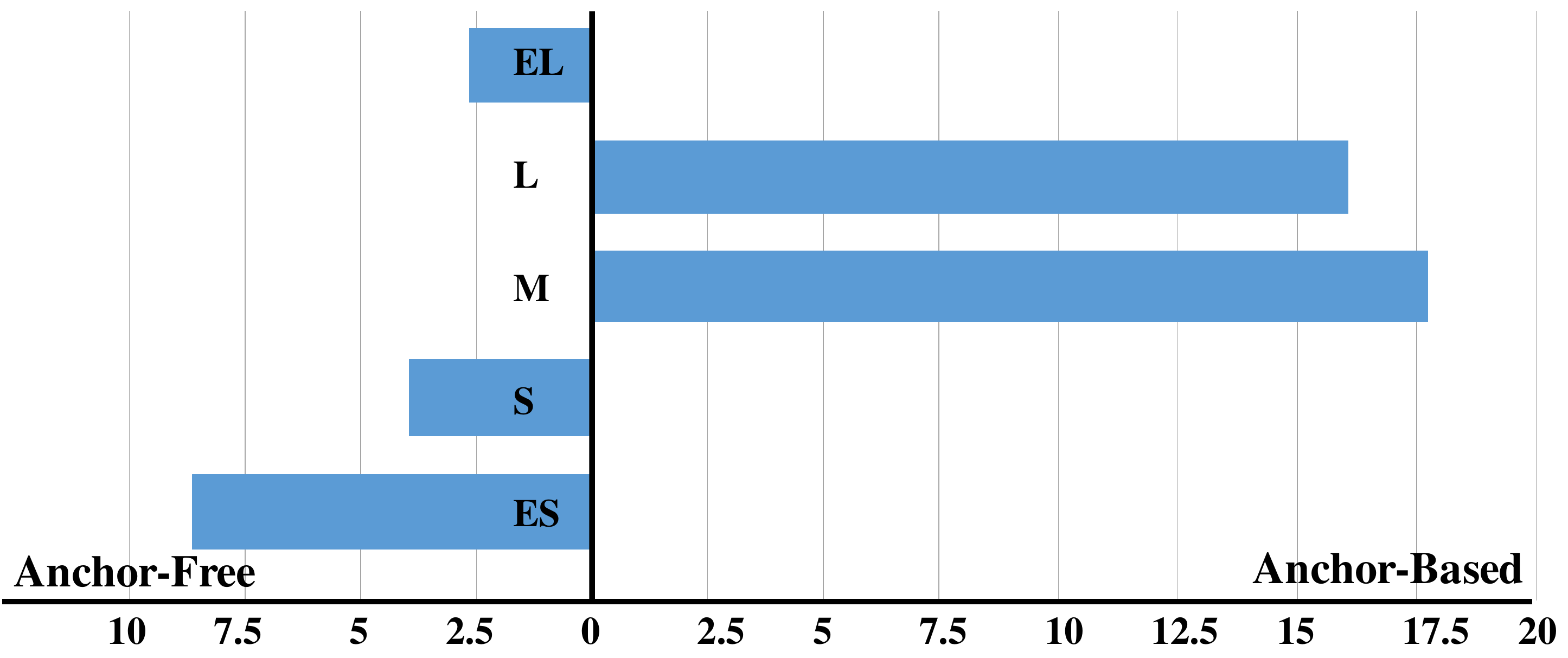}
  \caption{The performance superiority for anchor-free module (left) and anchor-based module (right) on each sub-dataset. Based on Table \ref{tab_subdataset_performance}, the horizontal axis stands for the absolute value of performance gap between the anchor-free module and the anchor-based module.}
  \label{fig1_motivation}
\vspace{-0.3cm}
\end{figure}

Recent years have witnessed increasing interests in temporal action localization. Early researches commonly adopt a detection-by-classification pipeline, such as S-CNN~\cite{shou2016temporal} and  Oneata et al. \cite{oneata2013action}. Recent works introduce the anchor mechanism into action localization, inspired by well-developed object detection researches \cite{liu2016ssd, girshick2015fast}. There are two kinds of action localization pipelines according to the anchor mechanism, namely the one-stage pipeline, such as SSAD~\cite{lin2017single}, SS-TAD~\cite{buch2017end} and GTAN~\cite{long2019gaussian}, and the two-stage pipeline, such as R-C3D~\cite{xu2017r}, TURN TAP~\cite{gao2017turn} and TAL\cite{chao2018rethinking}. The common mechanism of these approaches lies in representing action instances based on the pre-defined anchors, and the goal is to select the nearest anchor to the ground truth and predict regression parameters to adjust the anchor towards the ground truth.

The ingenious anchor mechanism and the classification-and-regression procedure are core factors for the success of anchor-based methods. For one thing, the pre-defined anchors are highly probable to cover the majority of action instances with large overlap ratios, especially when the model parameters are carefully designed and tuned so that the default anchors are consistent with the prior knowledge of the action instance distribution. For another thing, based on the reliable anchors, the regression procedure can refine the localization boundaries and precisely segment out action instances.

\begin{figure*}[htbp]
	\graphicspath{{figures/}}
	\centering
	\includegraphics[width=0.9\linewidth]{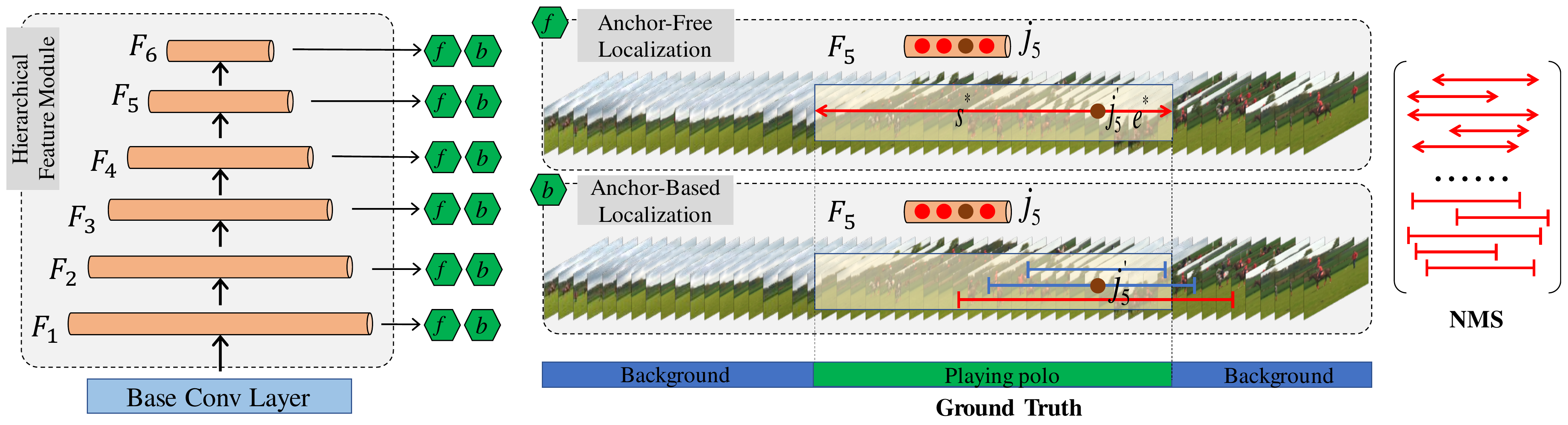}
	\caption{Framework of the proposed A2Net. The \emph{Base Conv Layer} converts raw features into inputs for the feature pyramid network. The feature pyramid network consists of $6$ cascaded 1D convolutional layers. Both anchor-free and anchor-based predictions are made at each hierarchical level. For temporal location $j_{5}$, A2Net maps it back onto the original video and obtains the corresponding location $j'_{5}$. The anchor-free localization module learns distances to the starting boundary $s^{*}$ and to the ending boundary $e^{*}$. The anchor-based localization module first matches the pre-defined anchors with ground truth (the matched anchor is shown in red), then it simultaneously learns classification score, overlap and regression parameters. In evaluation, the predictions from these two modules are merged together and then the standard NMS operation generates the final action localization results.}
	\label{fig2_framework}
\vspace{-0.3cm}
\end{figure*}

Although anchor-based pipelines have achieved exciting performance, some researchers already noticed that these methods may have one inherent limitation--they are not flexible enough~\cite{liu2019multi, chao2018rethinking}. The inflexibility lies in the pre-defined anchors. Precisely, the anchor-based mechanism set prior to the temporal location and duration of action instances, limiting its capability to dispose of action instances with various lengths. For example, TAL~\cite{chao2018rethinking} replaces a single convolutional layer with a collection of $K$ temporal ConvNets when making predictions. Each ConvNet is responsible for one kind of action instance with a specific duration, increasing the flexibility and achieving superior performance.

Inspired by the flexibility and comparable performance of anchor-free object detectors, e.g., CornerNet~\cite{law2018cornernet}, FCOS~\cite{tian2019fcos}, etc., we make an early effort to study the anchor-free mechanism and alleviate the restriction about action locations and durations caused by the pre-defined anchors. In this work, we propose representing an action instance as a point and its distances to the starting boundary and ending boundary, forming an anchor-free module for action localization. Compared with traditional anchor-based methods, the proposed anchor-free module shows superiority on flexibility. First of all, this module disposes each temporal location equally and sets no constrains for distances to action boundaries. Besides, every temporal point within the action scope can be responsible for detecting this action, increasing the number of training samples as well as the chance to accurately detect this action. In contrast, anchor-based counterparts only learn from the best-matched anchor.

We analyze the performance of the anchor-based module and the anchor-free module on THUMOS14 dataset, shown in Table \ref{tab_subdataset_performance} and Figure \ref{fig1_motivation} \footnote {The dataset is equally divided into $5$ sub-datasets according to the action duration $d$, including extremely short (ES) $d \in (0, 1.5s)$, short (S) $d \in [1.5s, 2.5s)$, medium (M) $d \in [2.5s, 4.2s)$, long (L) $d \in [4.2s, 6.9s)$, extremely long (EL) $d \in [6.9s, \propto]$.}. It can be found that the anchor-free module performs superior for \emph{extremely short} or \emph{extremely long} action instances, while the anchor-based module is more appropriate for action instances with \emph{medium} or \textit{long} length. Essentially, the reason for this phenomenon lies in the inherent flexibility of the anchor-free module and the stability of the classification-and-detection procedure adopted by the anchor-based module.

Considering the complementarity of the anchor-free module and the anchor-based module, we propose integrating these two modules, obtaining a simple but effective framework for action localization, namely A2Net. As shown in Figure \ref{fig2_framework}, A2Net cascades multiple 1D temporal convolutional layers together and builds a feature pyramid network, forming the backbone network. The anchor-free module simultaneously predicts the classification score and regresses the distances to the starting and ending boundaries. The anchor-based module first chooses the closest-matched anchor then refines the action boundaries via regression. These two modules share the same backbone and independently make predictions at each temporal location from every pyramid level. During inference, the predictions from both modules are collected together and the standard NMS generates the final action location results. We evaluate the proposed A2Net on two extensively used datasets, i.e. THUMOS14~\cite{THUMOS14} and ActivityNet v1.3~\cite{caba2015activitynet}.

Our contributions can be summarized as follows:

\begin{itemize}
  \item We make a comprehensive study on the anchor mechanisms for the action localization task and propose the novel A2Net by integrating the anchor-free mechanism and the traditional anchor-based mechanism in a simple but effective framework.
  \item We make an early exploration of the anchor-free action localization mechanism. The anchor-free module exhibits its flexibility for action localization tasks, especially for tackling extremely short and extremely long action instances.
  \item We verify the effectiveness of A2Net and the inherent complementarity between the anchor-free module and the anchor-based module on two benchmark datasets. Especially, on THUMOS14, A2Net achieves an mAP of $45.5$ at the threshold of $0.5$, demonstrating $2.7$ points gain over the previous strong competitor TAL \cite{chao2018rethinking} ($45.5\%$ versus $42.8\%$).
\end{itemize}

%%%%%%%%% Related work
\section{Related Work}

\subsection{Anchor-based Temporal Action Localization}
Existing anchor-based action localization methods can be categorized into two groups: one-stage approaches and two-stage approaches. SSAD~\cite{lin2017single} is an early exploration of one-stage action localization. It simultaneously predicts classification, overlapping and regression for each temporal location based on the default anchors. The hierarchical feature network captures actions with different durations from different hierarchical levels. Later, GTAN~\cite{long2019gaussian} proposes replacing the default anchor with a specific anchor, whose width can be dynamically learned. Meanwhile, MGG~\cite{liu2019multi} proposes a feature pyramid network and predicts action instances at each pyramid level.

Two-stage action localization approaches include R-C3D~\cite{xu2017r}, TURN TAP~\cite{gao2017turn}, CBR~\cite{gao2017cascaded},  TAL~\cite{chao2018rethinking}, etc. R-C3D extends the Faster R-CNN~\cite{ren2015faster} pipeline to temporal localization in the 1D sequence. In an end-to-end learning manner, it directly learns from frame clips, generates action proposals and classifies these proposals. Later, Gao et al.  propose considering the context information in TURN TAP~\cite{gao2017turn} and progressively regress the action boundary in CBR~\cite{gao2017cascaded}, under a two-stage anchor-based manner. Subsequently, TAL~\cite{chao2018rethinking} revisits anchor-based action localization pipelines and finds out the receptive filed is important for tackling actions with different scales.

The above traditional anchor-based methods severely rely on default anchors, which are pre-defined and fixed. For example, the anchor-based mechanism constrains the anchor point to be the center of the action instance. In contrast, the anchor-free mechanism can predict the distances to action boundaries autonomously. In essence, default anchors come from prior knowledge of action distribution. Thus, when prior knowledge is easy to obtain and reliable, the anchor-based method performs well. However, when the actions show drastic varieties and the prior knowledge is not stable, the flexibility of traditional anchor-based methods is limited. The proposed anchor-free module can flexibly tackle these actions, provides the complementarity to the anchor-based branch and improves the performance of A2Net.

\subsection{Actionness-guided Temporal Action Localization}
Anchor-based action localization methods adopt a top-down pipeline, first discover confident action segments, then refine the action boundaries. On the other hand, there are bottom-up pipelines for action localization \cite{shou2017cdc,  zhao2017temporal, lin2018bsn, lin2019bmn}, which utilize the frame-level actionness score to discover actions.

CDC~\cite{shou2017cdc} estimates frame-level actionness score via a Convolutional-De-Convolutional network. Then it cooperates with S-CNN~\cite{hou2017real} and uses the actionness score to refine action boundaries. Later, SSN~\cite{zhao2017temporal} calculates temporal actionness probability, discovers action proposals under different thresholds and groups the proposals to detect action instances. Subsequently, Lin et al. propose BSN~\cite{lin2018bsn}. It first evaluates actionness score, starting probability and ending probability via a temporal evaluation module. Then, BSN~\cite{lin2018bsn} enumerates potential proposals and estimates the confidence score for each proposal. BSN~\cite{lin2018bsn} is further developed by BMN~\cite{lin2019bmn}, which densely predicts the boundary matching confidence map and uses it to select action proposals. As BMN~\cite{lin2019bmn} pointed out, the actionness is a kind of useful guidance for precisely localizing boundary, but actionness-based methods are prone to detect false positive in the background video fragments. In the proposed A2Net framework, the anchor-free branch predicts a confidence score for each temporal point. It performs well for tackling actions with dramatic variety, providing complementarity for the anchor-based branch.

From the view of the anchor mechanism, the actionness-guided methods can be categorized into the anchor-free pipeline. However, both the high-level design and the detailed implementations are different between these actionness-guided methods and the proposed anchor-free module. These actionness-guided methods adopt a bottom-up pipeline and usually localize action instances via multiple separate procedures (e.g., BSN \cite{lin2018bsn}). In contrast, the proposed anchor-free module adopts a top-down pipeline and can discover action instances in a single forward pass.

Furthermore, there are some other action localization methods. S-CNN~\cite{shou2016temporal} classifies sliding-window proposals to localize action instances. CTAP~\cite{gao2018ctap} combines the sliding-windows based method (e.g., \cite{shou2016temporal}) and actionness-guided method (e.g.,~\cite{zhao2017temporal}) to discover actions. Liu et al. \cite{liu2018deep} use synthesized data to learn action representation. Besides, there are methods based on reinforcement learning~\cite{yeung2016end}, gated recurrent unit \cite{buch2017sst, zhu2017uncovering} graph convolutional network~\cite{zeng2019graph}. Some researches (e.g., \cite{zeng2019breaking}) localize action instances in the weakly supervised manner, which is beyond the scope of this paper.

\subsection{Object Detection using Anchor Mechanism}
The temporal action localization task is often inspired by object detection researches. Traditional object detectors can be categorized into two groups, anchor-based object detectors (e.g., SSD~\cite{liu2016ssd}, Fast RCNN~\cite{girshick2015fast}) and anchor-free object detectors (e.g., YOLO v1~\cite{redmon2016you}). As for anchor-based methods, object instances are represented via offset with respect to the pre-defined anchor boxes. As for anchor-free methods, object instances are directly learned via paired key points (e.g., CornerNet~\cite{law2018cornernet}) or distances to the boundaries (e.g., FCOS~\cite{tian2019fcos}, CenterNet~\cite{duan2019centernet}, etc.). Precisely, FCOS~\cite{tian2019fcos} predicts a 4D vector encoding the distance from the current point to the left, top, right, bottom boundaries. In the proposed A2Net,
the anchor-free module tries to learn the distances to action starting boundaries and ending boundaries. Although the high-level spirit of A2Net is similar with FCOS~\cite{tian2019fcos}, the implementation is thoroughly different (see Section \ref{subsection-af}) as A2Net is required to dispose of temporal frame sequences while FCOS~\cite{tian2019fcos} is designed for a single image. Furthermore, directly introduce the anchor-free mechanism into action localization cannot achieve good performance (see \ref{subsection-exp-ab}), it is the complementarity between anchor-free and anchor-based module that leads to the good performance of A2Net.

%------------------------------------------------------------------------
\section{Approach}

Given a video, A2Net aims to locate out all the action instance $B^{(i)}=\{t^{(i)}_{s}, t^{(i)}_{e}, c^{(i)}\}$ within this video, where $t^{(i)}_{s}$ and $t^{(i)}_{e}$ represent the starting time and ending time respectively, $c^{(i)}$ indicates the category label. As shown in Figure \ref{fig2_framework}, A2Net mainly contains $3$ modules, the \textit{Hierarchical Feature Module}, the \textit{Anchor-Free Localization Module}, and the \textit{Anchor-Based Localization Module}. First of all, A2Net cascades temporal 1D convolutional layers to construct the hierarchical feature network. Then, the anchor-free localization module predicts the classification scores and distances to action boundaries to determine the potential action instances. Meanwhile, the anchor-based localization module predicts the classification score, overlap value and regression parameters to determine the action instances. We elaborately present each module in this section.

\subsection{Feature Extraction}
We use I3D model \cite{carreira2017quo} to extract the features of video frames. Then, these features are processed by the \emph{Base Conv Layer} so as to expand the reception field and reduce the computation burden for the subsequent layers. As shown in Table \ref{tab_network}, there are two convolutional layers and a max-pooling layer for \emph{Base Conv Layer}. The kernel size for the first convolutional layer is $1$, being in charge of decreasing the channel size. The second convolutional layer fuses information in the temporal dimension.

\begin{table}[tbp]
  \caption{Network architecture for A2Net. K: kernel size, F: filter, S: stride, B: batch size, D: dimension for input feature, T: temporal length for input feature, t: temporal length for each prediction layer, C: category number.}
  \label{tab_network}
  \centering
  \small
  \begin{tabular}{c|ccc|c}
\toprule[2pt]
Name    &K              &F           &S            &Output Size \\
\hline
\multicolumn{5}{c}{Base Conv Layer} \\
\hline
input 	&None           &None		 &None	      &$B\times D \times T$   \\
base1	&1		        &512		 &1           &$B\times 512 \times T$ \\
base2	&9		        &512         &1           &$B\times 512 \times T$ \\
pool	&2              &None        &2           &$B\times 512 \times T/2$ \\
\hline
\multicolumn{5}{c}{Feature Module} \\
\hline
conv1 	&3              &512         &1           &$B\times 512 \times T/2$ \\
conv2 	&3              &1024        &2           &$B\times 1024 \times T/4$ \\
conv3 	&3              &1024        &2           &$B\times 1024 \times T/8$ \\
conv4 	&3              &2048        &2           &$B\times 2048 \times T/16$ \\
conv5 	&3              &2048        &2           &$B\times 2048 \times T/32$ \\
conv6 	&3              &4096        &2           &$B\times 4096 \times T/64$ \\
\hline
\multicolumn{5}{c}{Anchor-Free Localization Module} \\
%        &               &Anchor-Free Localization Module &         &            \\
\hline
AF conv1        &1              &512         &1           &$B\times 512 \times t$ \\
AF conv2        &3              &512         &1           &$B\times 512 \times t$ \\
AF conv3        &3              &512         &1           &$B\times 512 \times t$ \\
AF pred         &3              &C+2         &1           &$B\times (C+2) \times t$ \\
\hline
\multicolumn{5}{c}{Anchor-Based Localization Module} \\
%       &               &Anchor-Based Localization Module &         &            \\
\hline
AB pred         &3              &C+3         &1           &$B\times (C+3) \times t$ \\
\bottomrule[2pt]
  \end{tabular}
\vspace{-0.3cm}
\end{table}

\subsection{Hierarchical Feature Module}
\label{hie_feat_module}
The hierarchical feature module is in charge of generating features for each pyramid level. In implementation, both the recurrent network \cite{zhu2017uncovering} and the cascade of temporal convolutional layers are proper choices for hierarchical feature learning. Following previous works \cite{lin2017single, chao2018rethinking, long2019gaussian, lin2019bmn}, we adopt the temporal convolutions because of its implementational simplicity and fast training convergence.

Decreasing spatial scales and increasing channel number, is a common choice for the powerful modern ConvNets, such as VGG~\cite{simonyan2014very}, ResNet~\cite{he2016deep}, etc. As Han et al. ~\cite{han2017deep} pointed out, the increase of channel number is essential for the performance of ConvNets, because the diversity of high-level attributes increases along with the channel number increasing. GTAN \cite{long2019gaussian}, the current state-of-the-art action localization method on ActivityNet v1.3, adopts such a strategy. Following GTAN \cite{long2019gaussian}, we gradually increase the channel number along with the decrease of temporal length in the hierarchical feature module.

We adopt a six-level pyramid network in this module, generating features $\{F_{i}, i={1,2,...,6}\}$. Precisely, we use the convolution operation with stride $s_{i}=2$ to reduce the temporal length at each pyramid level. The network architecture is shown in Table \ref{tab_network}.

\subsection{Anchor-Based Localization Module}
The anchor-based localization module is prior-based and performs well for localizing action instances with medium and long duration. It aims to jointly predict classification scores $S^{ab}=\{s^{ab}_{0}, s^{ab}_{1}, ..., s^{ab}_{C}\}$, overlap value $p_{o}$ and regression parameters $\{\Delta_{c}, \Delta_{w}\}$.

In precise, there are $C$ categories for consideration. The overlap value $p_{o}$ estimates the IoU (Intersection over Union) between the predicted action scope and its closest ground truth. Based on the pre-defined anchor, an action instance can be represented via the regression parameters as follows:
\begin{equation}
\begin{aligned}
    c=&c^{d} + \alpha \Delta_{c} w^{d} \\
    w=&w^{d} \cdot exp(\beta \Delta_{w})
\end{aligned}
\label{eq_ab_regression}
\end{equation}
where $(c^{d}, w^{d})$ represents the center and width for the default anchor, $\alpha$ and $\beta$ are parameters to control the learning process.

Given a video sequence, the anchor-based localization module makes predictions for anchors at each temporal location. Only anchors with accurate predictions are used to learn about overlap value $p_{o}$ and the regression parameters $\{\Delta_{c}, \Delta_{w}\}$.

As a result of this, a necessary process for training is to match anchor predictions with ground truth. We calculate IoU between an anchor prediction and all ground truth instances. All anchors with $IoU > 0.5$ are considered as positive samples. The corresponding overlap target is set as $p^{*}_{o}=IoU$. During training, the ratio between positive and negative samples is $1$:$1$.

The anchor-based localization module is based on SSAD~\cite{lin2017single}. Our improvements lie in three aspects. (1) Feature. SSAD regards the classification score at the top fully-connected layer as a kind of feature. We argue that the low-dimension classification score is not descriptive enough and adopt the average output from the last max-pooling layer at I3D~\cite{carreira2017quo} model as a feature.  (2) Channel number for the feature network. SSAD adopts $512$ channels for all layers in the feature module. On the contrary, we gradually increase the channel number as discussed in subsection \ref{hie_feat_module}. (3) Network architecture. SSAD predicts action instances from $3$ layers. We argue that such a coarse prediction may leave out some short or long action instances, and we adopt a $6$-layer hierarchical feature module. The impact of these improvements is studied in the Experiment Section \ref{experiment}.

\subsection{Anchor-Free Localization Module}
\label{subsection-af}
\begin{figure}[tbp]
  \graphicspath{{figures/}}
  \centering
  \includegraphics[width=0.9\linewidth]{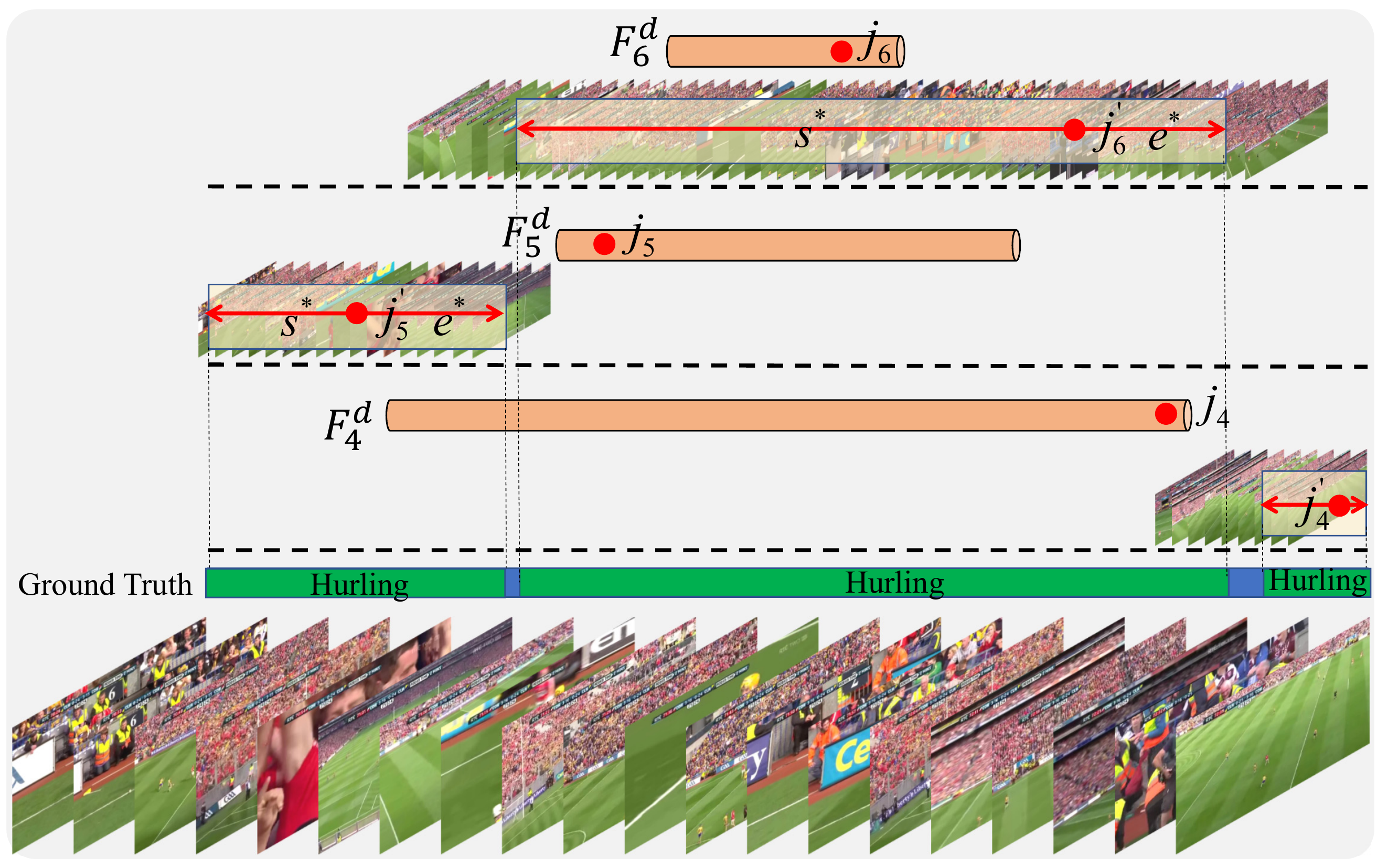}
  \caption{Illustration of anchor-free action localization. The example video contains $3$ action instances, which are assigned to different pyramid level according to their durations. This figure shows three pyramid levels, $F^{d}_{4}$, $F^{d}_{5}$ and $F^{d}_{6}$. Within each pyramid level, a temporal point is responsible for the assigned action instance, e.g., $j_{6}$ is responsible for the longest action instance. The temporal point can be mapped back to a location in the input length, e.g., $j^{'}_{6}$ is the corresponding location to $j_{6}$. The regression target is the distances from $j^{'}_{6}$ to the starting and ending boundary $(s^{*}, e^{*})$.}
  \label{fig3_afRegression}
\vspace{-0.3cm}
\end{figure}

The proposed anchor-free module regresses the distances from the center of an action instance to its boundaries. Compared with previous prediction methods over action proposals, anchor-free regression is more flexible and robust, especially for the action instances with extremely short or long durations.

Inspired by the hierarchical feature module \cite{lin2017single, long2019gaussian}, we divide action instances into several levels according to action temporal length. As shown in Figure \ref{fig3_afRegression}, the pyramid level $i$ only disposes of action instances with scale in range $[2^{i-1}, 2^{i}), i \in \{2, 3,..., 6\}$, for level $1$, the action scope is $[0, 2)$. For pyramid-level $i$ with temporal length $t$, the accumulated feature stride is $s_{i}$. Considering a temporal location $j, j \in \{0, 1, ..., t-1\}$ at feature map $F_{i}$, we can  map it back onto the input sequence:
\begin{equation}
    j'=\lfloor s_{i}/2 \rfloor + j\cdot s_{i}
\end{equation}
where $\lfloor * \rfloor$ indicates round down operation.
%$t$ is the temporal length for feature map $F_{i}$

After mapping back, if $j'$ falls into the action scope, we call it foreground point, otherwise background point. The anchor-free module predicts the classification score to determine whether a point falls into an action instance and its category label. The classification target for foreground and background point is $c^{*}=c^{i}$ and $c^{*}=0$ respectively. Only foreground points are utilized to regress the distance from current points to the action boundaries. The regression target can be calculated as
\begin{equation}
\begin{aligned}
    s^{*}=&j'-t^{(i)}_{s} \\
    e^{*}=&t^{(i)}_{e}-j'
\end{aligned}
\label{eq_regression}
\end{equation}

The anchor-free module uses two individual branches for predicting classification scores $S^{af}=(s^{af}_{0}, ..., s^{af}_{C})$ and regression distances to the starting boundary and ending boundary $(r_{s}, r_{e})$. The network architecture is reported in Table \ref{tab_network}.

\subsection{Training}
The A2Net simultaneously makes predictions from anchor-free localization module and anchor-based localization module. The former module generates the classification loss $L^{af}_{cls}$ and regression loss $L^{af}_{reg}$, while the latter module generates the classification loss $L^{ab}_{cls}$, overlap loss $L^{ab}_{o}$ and regression loss $L^{ab}_{reg}$. We combine the losses from the anchor-free module $L^{af}$ and the losses from the anchor-based module $L^{ab}$ to train A2Net in an end-to-end manner.

\begin{equation}
    L=L^{af} + \gamma L^{ab}
    \label{eq_total_loss}
\end{equation}
where $\gamma$ is the coefficient to balance these two kinds of losses. $L^{af}$ and $L^{ab}$ can be calculated as follows:

\begin{equation}
\begin{aligned}
    L^{af}=&L^{af}_{reg} + \gamma^{af}L^{af}_{cls} \\
    L^{ab}=&L^{ab}_{cls} + \gamma^{ab}_{1}L^{ab}_{o} + \gamma^{ab}_{2}L^{ab}_{reg}
    \label{eq_af_ab_loss}
\end{aligned}
\end{equation}
where $\gamma^{af}$, $\gamma^{ab}_{1}$ and $\gamma^{ab}_{2}$ are coefficients to balance different losses.

\noindent
\textbf{Anchor-Free Classification Loss}. Based on the classification scores $s^{af}=(s^{af}_{0}, ..., s^{af}_{C})$, A2Net calculates the anchor-free classification loss among all $N$ temporal locations by the standard cross-entropy loss.
\begin{equation}
    L^{af}_{cls} = -\frac{1}{N}\sum^{N}_{n=1}\sum^{C}_{i=0} log(\frac{exp(s^{af}_{i})}{\sum^{C}_{j=0}exp(s^{af}_{j})})
\end{equation}

\noindent
\textbf{Anchor-Free Regression Loss}. For a temporal point falling into the action scope, the anchor-free localization module regresses the distances to the starting and ending boundaries $(r_{s}, r_{e})$. The regression loss is calculated among $N^{af}_{p}$ foreground points and adopts the smooth L1 loss~\cite{girshick2015fast}.
\begin{equation}
\begin{aligned}
    L^{af}_{reg}=\frac{1}{N^{af}_{p}} \sum^{N^{af}_{p}}_{n=1} (&smoothL_{1}(r_{s}, s^{*}) \\
                 &+ smoothL_{1}(e_{s}, e^{*}))
\end{aligned}
\end{equation}

\noindent
\textbf{Anchor-Based Classification Loss}. The anchor-based module predicts classification scores $S^{ab}=\{s^{ab}_{0}, s^{ab}_{1}, ..., s^{ab}_{C}\}$ for all $N$ temporal locations, among which we select $N^{ab}_{p}$ positive anchors and $N^{ab}_{n}$ negative ones to calculate classification loss via cross-entropy loss.
\begin{equation}
    L^{ab}_{cls} = -\frac{1}{N^{ab}_{p}+N^{ab}_{n}}\sum^{N^{ab}_{p}+N^{ab}_{n}}_{n=1}\sum^{C}_{i=0} log(\frac{exp(s^{ab}_{i})}{\sum^{C}_{j=0}exp(s^{ab}_{j})})
\end{equation}

\noindent
\textbf{Anchor-Based Overlap Loss}. The overlap $p_{o}$ is the estimation of IoU between the predicted action instances and its closest ground truth, calculated among $N^{ab}_{p}$ anchors. We use the mean square error loss.
\begin{equation}
    L^{ab}_{o} = \frac{1}{N^{ab}_{p}} \sum^{N^{ab}_{p}}_{n=1} (p_{o} - p^{*}_{o})^{2}
\end{equation}

\noindent
\textbf{Anchor-Based Regression Loss}. Given regression target $(\Delta^{*}_{c}, \Delta^{*}_{w})$ and the prediction $(\Delta_{c}, \Delta_{w})$ for $N^{ab}_{p}$ selected anchors, the anchor-based regression loss can be calculated as follows:
\begin{equation}
\begin{aligned}
    L^{ab}_{reg} = \frac{1}{N^{ab}_{p}}\sum^{N^{ab}_{p}}_{n=1} (&smoothL_{1}(\Delta_{c}, \Delta^{*}_{c}) \\
                   &+ smoothL_{1}(\Delta_{w}, \Delta^{*}_{w}))
\end{aligned}
\end{equation}

\begin{figure}[htbp]
	\graphicspath{{figures/}}
	\centering
	\includegraphics[width=0.9\linewidth]{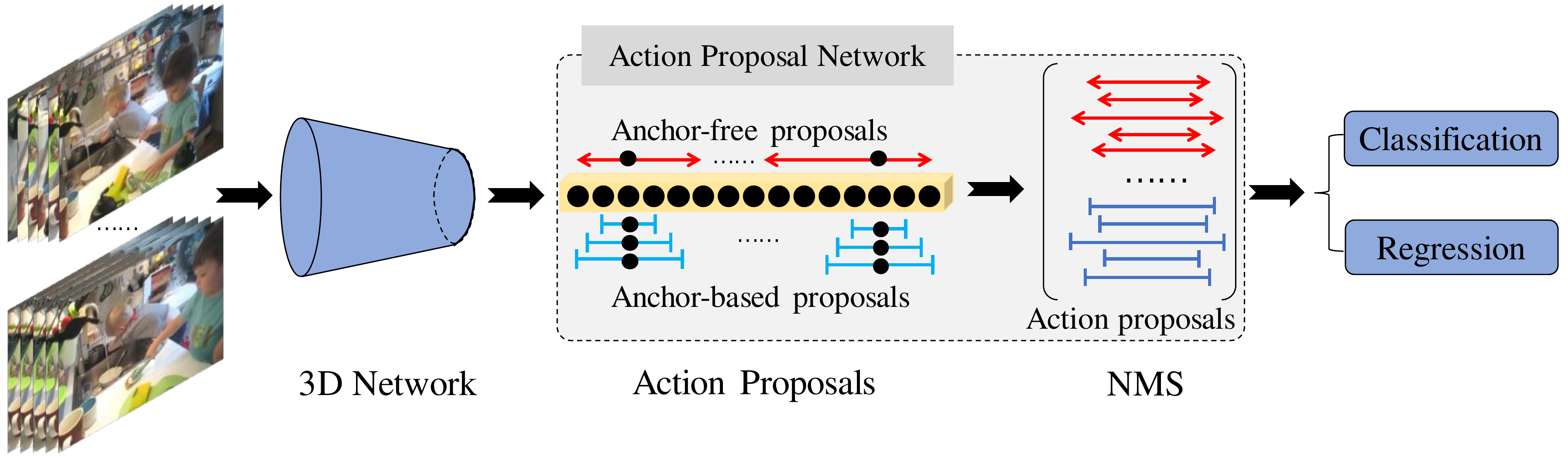}
	\caption{Illustration of the anchor-free module to facilitate the proposal generation. The anchor-free module is inserted into the \textit{Action Proposal Network}, generating anchor-free proposals in parallel with the conventional anchor-based proposals. These proposals are gathered together for NMS and other subsequent procedures.}
	\label{fig5_af_proposal}
\vspace{-0.3cm}
\end{figure}

\subsection{Inference}
During inference, we forward the feature sequence through the network and obtain the predictions from the anchor-free module and the anchor-based module. As for the anchor-free module, the predicted action boundaries $(s^{af}, e^{af})$ can be obtained via inverting equation~\ref{eq_regression}. The maximum value of classification score $S^{af}$ is regarded as the confidence for the localization results. According to the anchor-based module, we can obtain the predicted action instances via equation~\ref{eq_ab_regression}, along with classification scores $S^{ab}$ and overlap values $p_{o}$. We fuse $S^{ab}$ and $p_{o}$ to represent the confidence score $s^{ab}_{conf}$ for the anchor-based detection:
\begin{equation}
s^{ab}_{conf}=p_{o} \times max(S^{ab})
\label{eq_fuse_score}
\end{equation}
Finally, action localization results for the anchor-free and anchor-based modules are merged together. Specifically, as for each point at a hierarchical level, we can obtain two kinds of action localization results: the anchor-free localization results and the anchor-based localization results. We collect the localization results from all points together, perform NMS and obtain the final localization results.

\subsection{Anchor-free Module for Generating Action Proposals}
\label{af_proposal}

The proposed A2Net integrates the anchor-based mechanism and the anchor-free mechanism into a unified framework. Although the anchor-based module is adapted from the one-stage action localization method SSAD \cite{lin2017single}, we believe that the complementarity between the anchor-based module and the anchor-free module also applies to two-stage action localization methods.

In the two-stage classification-and-regression pipeline, the \textit{Action Proposal Network} is responsible for generating high-quality proposals, which are the foundation for the subsequent classification and regression process. As shown in Figure \ref{fig5_af_proposal}, we introduce the anchor-free mechanism into the \textit{Action Proposal Network}. The anchor-free module generates proposals in parallel with the traditional anchor-based module. All proposals are gathered together and then perform NMS to generate candidate proposals. We show experimental results in \ref{subsection-af-proposal}.

\section{Experiments}
\label{experiment}

\subsection{Experiment Setup}
\noindent
\textbf{Dataset}
We carry on experiments on two widely used benchmarks THUMOS14~\cite{THUMOS14} and ActivityNet v1.3~\cite{caba2015activitynet}. THUMOS14 is made for action recognition and localization. The action localization part contains $20$ labeled categories, including $213$ untrimmed videos in the validation set and $200$ untrimmed videos in the test set. Following former researches~\cite{lin2018bsn, chao2018rethinking, lin2017single} etc., we learn from the validation set and evaluate the performance on the test set. ActivityNet v1.3 contains $19994$ videos, $200$ labeled categories. This dataset comprises of training, validation and testing dataset with the number ratio of $2$:$1$:$1$. Challenges for ActivityNet v1.3 come from the large variety of action scales, intra-category differences and inter-category similarity, etc.

\noindent
\textbf{Metric}
We adopt the official evaluation metric mAP on both THUMOS14~\cite{THUMOS14} and ActivityNet v1.3~\cite{caba2015activitynet} datasets. Given a threshold, the interpolated average precision (AP) is first calculated and the average of APs generates mAP. On THUMOS14, the threshold is set as $0.1$:$0.1$:$0.7$. The performance comparison focuses on the score under the threshold $0.5$. On ActivityNet v1.3, we report the performance under the threshold $\{0.50, 0.75, 0.95\}$ as well as the average mAP under the threshold $0.5$:$0.05$:$0.95$.

\noindent
\textbf{Feature Extraction}
We extract I3D~\cite{carreira2017quo} feature from raw video data to represent video sequences. Raw videos are first decomposed into frames with frame rate $30fps$ for both THUMOS14 and AcitivityNet v1.3 datasets. Then the TV-L1~\cite{zach2007duality}  algorithm is used to estimate the optical flow. Finally, I3D model extracts features from stacked consecutive $16$ frames from both the frames and the optical flow data. The feature sequence can be represented as $f \in R^{D\times T}$, where $D$ and $T$ indicate the feature dimension and the number of feature vectors, respectively.

\subsection{Implementation Details}
For experiments on THUMOS14, we generate sliding windows for training and evaluation, and each sliding window contains $512$ frames. For experiments on ActivityNet v1.3, we use one sliding window for each video. The feature sequence is scaled into fixed temporal length (i.e., $128$) via linear interpolation.

As for action instance representation in equation \ref{eq_ab_regression}, we set $\alpha = 0.0001$ and $\beta = 0.0001$. In equation \ref{eq_total_loss}, $\gamma$ is set as $1$. For calculating loss in equation \ref{eq_af_ab_loss}, the trade-off parameters are set as $\gamma^{af}=30$, $\gamma^{ab}_{1}=10$, $\gamma^{ab}_{2}=10$. We use Adam~\cite{kingma2014adam} to optimize the network. The batch size is set as $32$. The initial learning rate is $1e-4$.
On THUMOS14, the total epoch number is $40$ and the learning rate is decayed by a factor $0.1$ at epoch $30$. On ActivityNet v1.3, the total epoch number is $24$ and the learning rate is decayed by a factor $0.1$ at epoch $16$.

The proposed A2Net is implemented based on PyTorch 1.4 \cite{paszke2019pytorch}. We perform experiments with one NVIDIA TITAN Xp GPU, Intel Xeon E5-2683 v3 CPU and 128G memory.

\subsection{Ablation studies}
\label{subsection-exp-ab}

\noindent
\textbf{The complementary performance between the anchor-free module and the anchor-based module}. In order to verify the performance complementarity between the anchor-free module and the anchor-based module, we utilize each individual module to localize action instances. The results are reported in the first two rows of Table \ref{tab_complenmentarity_thumos}. Besides, we examine the effect of directly fusing the localization results from the anchor-free module and the anchor-based module. In precise, we individually train two action localization models: one only uses the anchor-free module and the other only uses the anchor-based module. As for each point, we can obtain two kinds of action localization results, the anchor-free localization results, and the anchor-based localization results. We collect these localization results from all temporal points together, perform NMS and obtain the final action localization results, shown in the third row of Table \ref{tab_complenmentarity_thumos}.

\begin{table}[tbp]
	\caption{Complementary performance for anchor-free (AF) module and anchor-based (AB) module on THUMOS14 dataset, measured by mAP. }
	\label{tab_complenmentarity_thumos}
	\centering
	\small
	\begin{tabular}{c|ccccccc}
		\toprule[2pt]
		tIoU  &0.1       &0.2       &0.3       &0.4       &0.5       &0.6       &0.7    \\
		\midrule[1pt]
		AF    &52.1      &51.2      &49.2      &44.2      &36.6      &26.0      &15.0   \\
		AB 	  &60.5      &59.1      &56.4      &50.1      &39.7      &25.9      &12.8  \\
		fuse  &60.7      &59.7      &57.5      &52.0      &42.1      &29.4      &15.7  \\
		A2Net &\textbf{61.1}	     &\textbf{60.2}	    &\textbf{58.6}	   &\textbf{54.1}	  &\textbf{45.5}	     &\textbf{32.5}	    &\textbf{17.2}  \\
		\bottomrule[2pt]
	\end{tabular}
\vspace{-0.3cm}
\end{table}

From Table \ref{tab_complenmentarity_thumos}, we can find that, although the anchor-based module itself can achieve good performance, A2Net performs better with the combination of the anchor-based module and the anchor-free module, especially under high IoU. Under the metric mAP@$0.5$, the $4.0$ points performance gap between anchor-based module and A2Net reflects the complementarity between the anchor-free module and anchor-based module. The reason for such performance complementarity lies in the distinct but complementary feature patterns for the anchor-free module and the anchor-based module. The anchor-based module is designed to learn the offsets with respect to the default anchors. Thus, features are sensitive to relative distances. On the contrary, the anchor-free module is designed to learn the distances to action boundaries, which is sensitive to absolute distances and more flexible.

Furthermore, directly fusing the localization results from the anchor-free method and the anchor-based method cannot reach high performance. Specifically, \emph{fuse} falls behind A2Net with a margin of $3.4$ points. This demonstrates the necessity of jointly learning the anchor-free module and the anchor-based module, rather than just fusing the localization results.

From Table \ref{tab_complenmentarity_thumos}, we notice that the performance for the individual anchor-free method is lower than the performance of the anchor-based method under threshold $0.1$:$0.1$:$0.5$, and analyze the reason for this. We collect the top $200$ proposals from both the anchor-free method and the anchor-based method, then calculate the recall rate. The anchor-based method can recall $73.75\%$ proposals, and the anchor-free method can recall $74.87\%$ proposals. This suggests that the anchor-free module is capable to generate high-quality action proposals. It is the limited classification ability that constrains the action localization performance. Similar problems occur for anchor-free object detectors. For instance, FCOS \cite{tian2019fcos} uses an extra center-ness branch to assist classification. Designing an extra branch to assist the classification may boost the performance of the anchor-free module, but this will take the anchor-free module complicated. This research focuses on the complementarity between the anchor-free module and the anchor-based module and thus we choose the current anchor-free module.

\begin{table}[tbp]
	\caption{Complementarity of anchor-free (AF) module and anchor-based (AB) module on ActivityNet v1.3 dataset, measured by mAP.}
	\label{tab_complenmentarity_anet}
	\centering
	\small
	\begin{tabular}{c|cccc}
		\toprule[2pt]
		&0.50	&0.75	&0.95	&mAP \\
		\midrule[1pt]
		anchor free 	    &41.89		&25.96		&3.36		&25.97	 \\
		anchor based	    &42.96		&26.88		&3.72		&25.68	 \\
		fuse 	&40.61      &28.50		&\textbf{3.92}		&26.40   \\
		A2Net 	&\textbf{43.55}		&\textbf{28.69}		&3.70       &\textbf{27.75}   \\
		\bottomrule[2pt]
	\end{tabular}
\vspace{-0.3cm}
\end{table}

Table \ref{tab_complenmentarity_anet} reports the performance on ActivityNet v1.3 dataset. The anchor-free module and the anchor-based module achieve similar average mAP, $25.97$ and $25.68$ respectively. Directly fusing the prediction results can improve the average mAP to $26.40$. By simultaneously learning the anchor-free module and the anchor-based module, A2Net can boost the performance to $27.75$, verifying the complementarity between the anchor-free module and the anchor-based module. In Table \ref{tab_complenmentarity_anet}, we can notice that the improvement under threshold $0.95$ is not obvious. The main reason is that mAP under threshold $0.95$ is a strict criterion. Thus, the performance is usually low and the improvement of A2Net is not obvious. Under threshold $0.95$, the anchor-based method and the fuse method performs better than A2Net. The reason is that the anchor-based method is guided by the prior knowledge and is more potential to precisely localize boundaries for action instances with medium length. However, considering mAP under threshold $0.50$:$0.05$:$0.95$, we can find that A2Net performs superior.

\begin{table}[tbp]
\vspace{-0.15cm}
  \centering
  \caption{Ablation analysis about the merge strategy. We perform experiments on THUMOS14 dataset and report mAP under threshold $0.5$.}
    \begin{tabular}{c|ccccc}
    \toprule[2pt]
    $\lambda$ & 0.2   & 0.4   & 0.5   & 0.6   & 0.8 \\
    \midrule[1pt]
    mAP   & 44.1  & 44.6  & \textbf{45.5}  & 45.2  & 44.2 \\
    \bottomrule[2pt]
    \end{tabular}%
  \label{tab:ablation-merge}%
\vspace{-0.15cm}
\end{table}%

In inference, A2Net merges the localization results from the anchor-free branch and anchor-based branch. Since the quality of these two kinds of localization results may be different, we adjust the confidence score with a parameter $\lambda$ when merging. Specifically, the confidence score is multiplied with a coefficient $\lambda$ for localization results from the anchor-based branch, while it is multiplied with a coefficient $1 - \lambda$ for results from the anchor-free branch. As shown in Table \ref{tab:ablation-merge}, A2Net achieves the best performance $45.5$ when $\lambda = 0.5$. This indicates the localization results from these two branches show similar quality.

\noindent
\textbf{Hyper-parameter and Loss curves}. As shown in equation (\ref{eq_total_loss}) and equation (\ref{eq_af_ab_loss}), the complete loss function contains 4 coefficients. Here we discuss how to set these hyper-parameters and their impacts. As for the anchor-based loss $L^{ab}$, $\gamma^{ab}_{1}$ and $\gamma^{ab}_{2}$ work together to balance the impacts of the classification loss $L^{ab}_{cls}$, the overlap loss $L^{ab}_{o}$ and the regression loss $L^{ab}_{reg}$. We follow the anchor-based method SSAD \cite{lin2017single} to set $\gamma^{ab}_{1}=10$ and $\gamma^{ab}_{2}=10$. As for the anchor-free loss $L^{af}$, the regression loss $L^{af}_{reg}$ uses the smooth L1 loss, while the classification loss $L^{af}_{cls}$ uses the cross-entropy loss. The magnitudes of these two losses are quite different, i.e., the classification loss is much small. We set $\gamma^{af}=30$ so as to make $L^{af}_{reg}$ and $L^{af}_{cls}$ in a similar magnitude. As shown in Fig. 6 (a), with the help of $\gamma^{af}$, the enhanced classification loss shows a similar magnitude with the regression loss.
\begin{figure}[tbp]
	\graphicspath{{figures/}}
	\centering
	\includegraphics[width=0.9\linewidth]{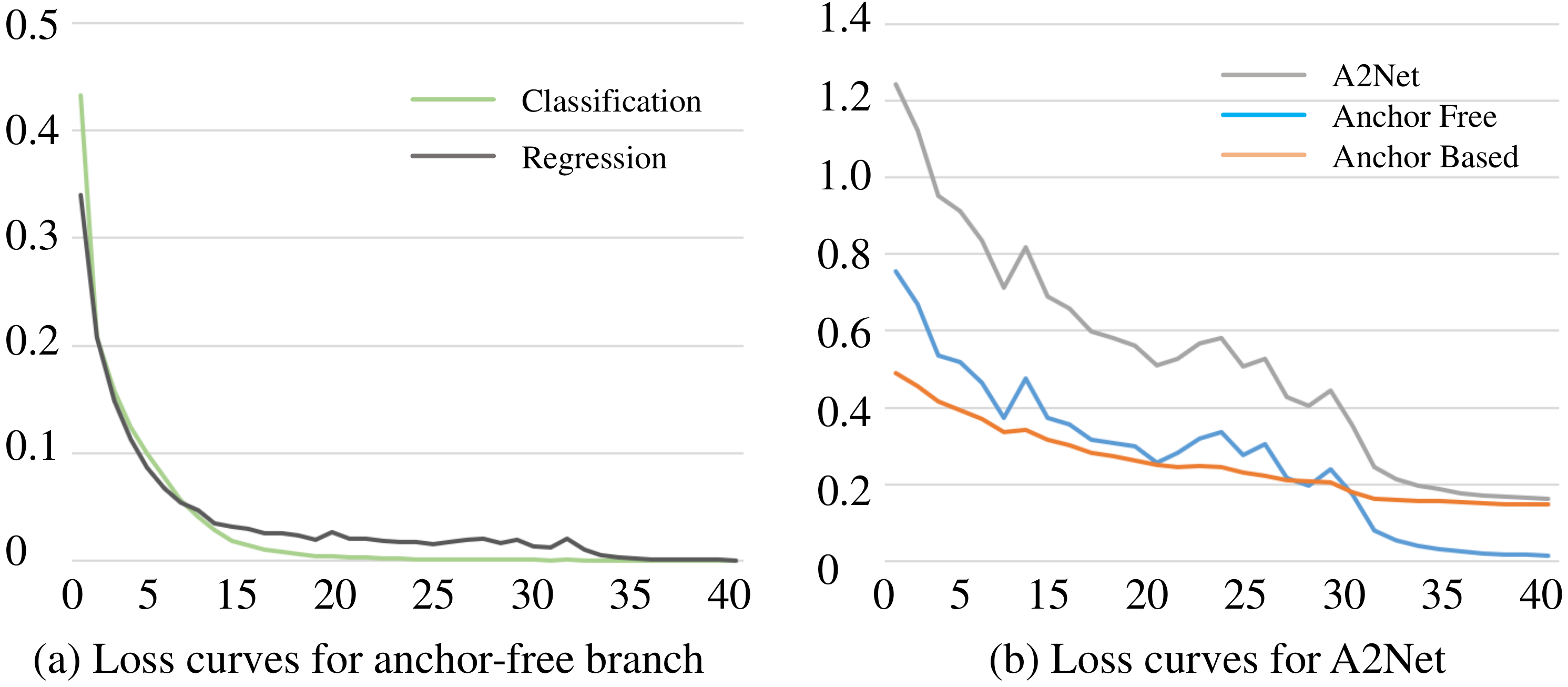}
	\caption{Loss curves of anchor-free branch and A2Net.}
	\label{fig_loss_curve}
    \vspace{-0.2cm}
\end{figure}
The complete loss simultaneously considers the loss from the anchor-free branch $L^{af}$ and the loss from the anchor-based branch $L^{ab}$. We treat these two losses equally and set $\gamma=1$. In Fig. 6 (b), we show loss curves for the anchor-free branch, the anchor-based branch and A2Net. Within $40$ epochs, the loss curves gradually decrease and converge to a small value.

\begin{table}[bp]
	\centering
    \vspace{-0.3cm}
	\caption{Temporal length and computational costs (measured by FLOPs) for different hierarchical levels.}
	\begin{tabular}{c|cc}
    \toprule[2pt]
    Level & Temporal length for each layer & FLOPs/G \\
    \midrule
    3     & 16-8-4 & 1.02 \\
    4     & 32-16-8-4 & 1.11 \\
    5     & 32-16-8-4-2 & 1.18 \\
    6     & 64-32-16-8-4-2 & 1.24 \\
    \bottomrule[2pt]
    \end{tabular}%
	\label{tab:tem-length-pyramid}%
\end{table}%

\noindent
\textbf{The influence of network architecture}. Although Table \ref{tab_complenmentarity_thumos} and Table \ref{tab_complenmentarity_anet} demonstrate the complementarity between the anchor-free module and the anchor-based module. Someone may argue that it is the increased parameters (e.g., more layers and more channels) that lead to the effective performance of A2Net. We carry on three experiments with A2Net, only anchor-free module and only anchor-based module under different layers and channels. In precise, A2Net contains $6$ pyramid levels, and each level is utilized to make predictions. We streamline the number of pyramid levels to $3, 4, 5$ and verify the performance of the individual anchor-free module, the individual anchor-based module, and A2Net. The detailed temporal length of each hierarchical level is shown in Table \ref{tab:tem-length-pyramid}. In precise, for a $3$-level network, temporal length for each layer is "16-8-4", indicating that there are $3$ hierarchical feature layers and the temporal length is $16$, $8$ and $4$, respectively.

\begin{figure}[tbp]
	\graphicspath{{figures/}}
	\centering
	\includegraphics[width=0.9\linewidth]{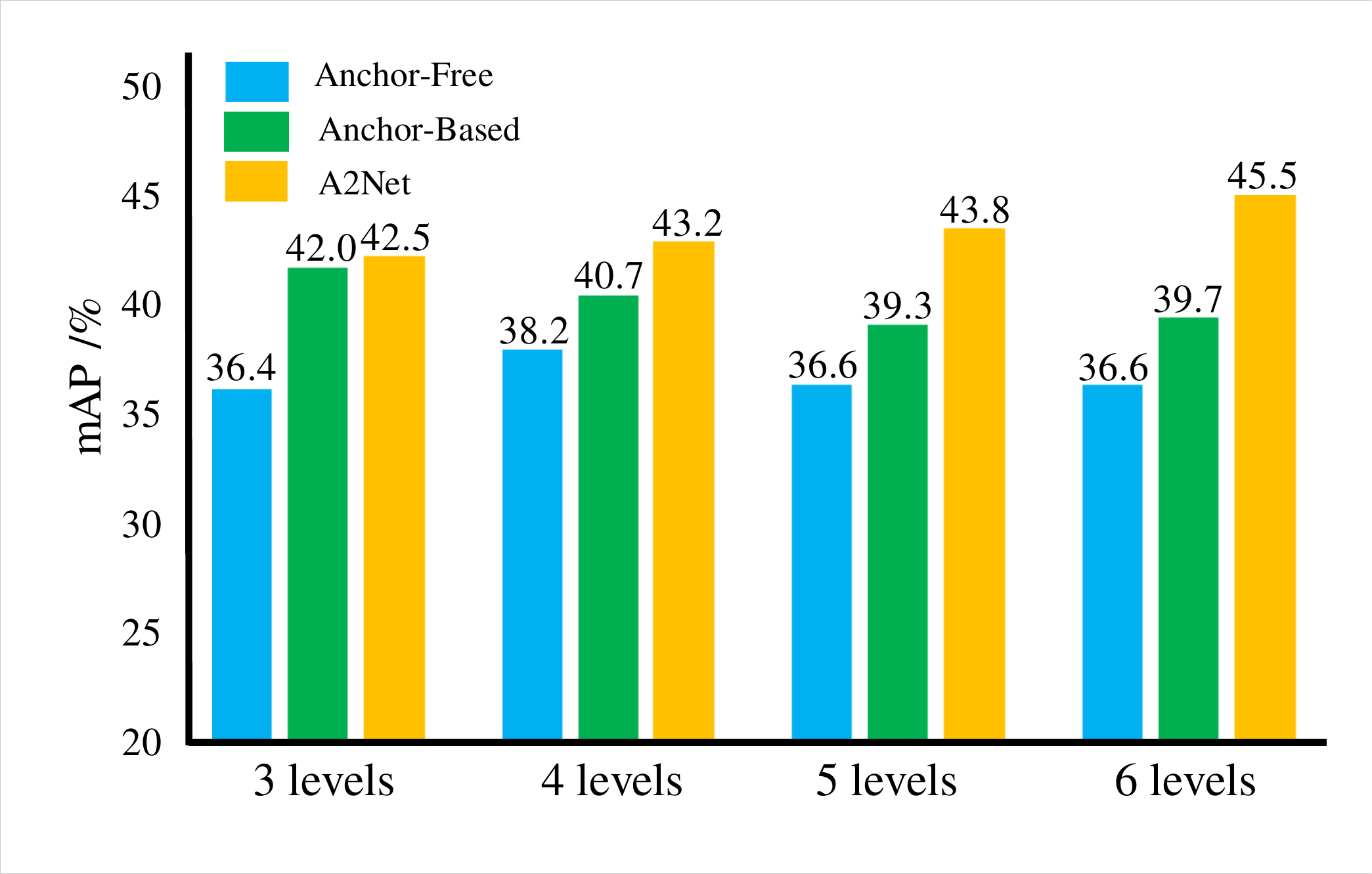}
	\caption{Performance for anchor-free module, anchor-based module and A2Net under different pyramid levels.}
	\label{fig5_perf_layers}
    \vspace{-0.3cm}
\end{figure}

Figure \ref{fig5_perf_layers} reports the experiment results. First of all, the individual anchor-free module or anchor-based module only witnesses limited improvement or even fluctuation. Besides, the performance of A2Net gradually improves along with the increase of the pyramid level, from $42.5$ to $45.5$. A2Net gradually increases the channel number along with the increase of the layer number, in pursuit of a good trade-off between the performance and the computational cost. Furthermore, we equip A2Net with more channels but only obtain slight performance improvement, which indicates the current network architecture is a good choice.

Essentially, there are two reasons for these experimental results. For one thing, it is the false positives that limit the performance of the anchor-free method and the anchor-based method. Given a $n$-level network, adding one extra layer will increase many temporal points. For example, a $6$-level network has $2^{n+1}$ more temporal points than a $5$-level network. The increase of temporal points requires increased predictions from the network. However, neither the anchor-free method nor the anchor-based method can individually make full use of the increased capability of more levels. Then, many false positives occur in the localization results, (e.g., the localization result contains many short action segments, which are wrong localization results), and the performance fluctuates. Given another piece of evidence, GTAN \cite{long2019gaussian} shows similar experiment results. The baseline of GTAN \cite{long2019gaussian} is similar to the anchor-based branch of the proposed A2Net, but contains more layers and channels. Its performance is mAP@$0.5$=$33.5$, inferior to mAP@$0.5$=$36.6$ achieved by the anchor-based method with 6 levels.

For another thing, A2Net jointly learns the anchor-free module and the anchor-based module. Because these two modules are complementary with respect to the learned feature patterns, A2Net can sufficiently utilize the enlarged capacity for the increase of the layer number and the channel number. Consequently, the performance of A2Net gradually increases when the pyramid level increases from $3$ to $6$. In summary, it is the complementarity between anchor-free and anchor-based module that guarantees the good performance of A2Net, rather than the increased layer number or channel number.

Furthermore, we construct a $7$-level architecture for A2Net but do not see noticeable performance gains. Thus, we select the $6$-level architecture for subsequent experiments.

\noindent
\textbf{The influences of features and fusion manners}.
There are some influential settings for action localization framework, e.g. features, predictions fusing manners. We study the impact of these factors.
\begin{table}[tbp]
	\caption{The influences of features with different modalities and the fusion manner. We measure the performance by mAP under different thresholds on the THUMOS14 dataset. Spa: only use spatial feature; Tem: only use temporal feature; LFuse: late fusion.}
	\label{tab_abl_feature}
	\centering
	\small
	\begin{tabular}{c|ccccccc}
		\toprule[2pt]
		&0.1       &0.2       &0.3       &0.4       &0.5       &0.6       &0.7    \\
		\midrule[1pt]
		Spa     &49.2      &47.9      &45.0      &40.5      &31.3      &19.9      &10.0   \\
		Tem     &57.0      &56.4      &54.4      &50.2      &41.1      &29.5      &17.6  \\
		LFuse	&60.0      &59.1      &57.0      &52.3      &43.5      &31.1      &\textbf{18.4}  \\
		\midrule[1pt]
		A2Net	&\textbf{61.1}	   &\textbf{60.2}	  &\textbf{58.6}	     &\textbf{54.1}	    &\textbf{45.5}	   &\textbf{32.5}	  &17.2	 \\
		\bottomrule[2pt]
	\end{tabular}
\vspace{-0.3cm}
\end{table}

\begin{table*}[htbp]
  \centering
  \caption{Comparison with state-of-the-art methods on the THUMOS14 test dataset, measured by mAP.}
    \begin{tabular}{cc|c|c|c|ccccccc|c}
    \toprule[2pt]
    \multicolumn{2}{c|}{} & Research & Publication & Feature & 0.1   & 0.2   & 0.3   & 0.4   & 0.5   & 0.6   & 0.7   & avg-mAP (0.1:0.5) \\
    \midrule
    \multicolumn{2}{c|}{\multirow{8}[2]{*}{Early works}} & SLM~\cite{richard2016temporal} & CVPR2016 & iDT   & 39.7  & 35.7  & 30.0  & 23.2  & 15.2  & --    & --    & 28.8  \\
    \multicolumn{2}{c|}{} & Yeungelal.~\cite{yeung2016end} & CVPR2016 & --    & 48.9  & 44.0  & 36.0  & 26.4  & 17.1  & --    & --    & 26.4  \\
    \multicolumn{2}{c|}{} & Yuanelal.~\cite{yuan2017temporal} & CVPR2017 & TS    & 51.0  & 45.2  & 36.5  & 27.8  & 17.8  & --    & --    & 35.7  \\
    \multicolumn{2}{c|}{} & Yuanelal.~\cite{yuan2016temporal} & CVPR2016 & PSDF  & 51.4  & 42.6  & 33.6  & 26.1  & 18.8  & --    & --    & 34.5  \\
    \multicolumn{2}{c|}{} & S-CNN~\cite{shou2016temporal} & CVPR2016 & --    & 47.7  & 43.5  & 36.3  & 28.7  & 19.0  & 10.3  & 5.3   & 35.0  \\
    \multicolumn{2}{c|}{} & Houetal.~\cite{hou2017real} & BMVC2017 & iDT   & 51.3  & --    & 43.7  & --    & 22.0  & --    & --    & -- \\
    \multicolumn{2}{c|}{} & SST~\cite{buch2017sst} & CVPR2017 & C3D   & --    & --    & 37.8  & --    & 23.0  & --    & --    & -- \\
    \multicolumn{2}{c|}{} & SS-TAD~\cite{buch2017end} & BMVC2017 & C3D   & --    & --    & 45.7  & --    & 29.2  & --    & 9.6   & -- \\
    \midrule
    \multicolumn{1}{c|}{\multirow{7}[4]{*}{Anchor}} & \multirow{2}[2]{*}{one stage} & SSAD~\cite{lin2017single} & ACM2017 & TS    & 50.1  & 47.8  & 43.0  & 35.0  & 24.6  & --    & --    & 40.1  \\
    \multicolumn{1}{c|}{} &       & GTAN~\cite{long2019gaussian} & CVPR2019 & P3D   & \textbf{69.1 } & \textbf{63.7 } & 57.8  & 47.2  & 38.8  & --    & --    & 55.3  \\
\cmidrule{2-13}    \multicolumn{1}{c|}{} & \multirow{5}[2]{*}{two stage} & TURN~\cite{gao2017turn} & ICCV2017 & C3D   & 54.0  & 50.9  & 44.1  & 34.9  & 25.6  & --    & --    & 41.9  \\
    \multicolumn{1}{c|}{} &       & Daietal.~\cite{dai2017temporal} & ICCV2017 & TS    & --    & --    & --    & 33.3  & 25.6  & 15.9  & 9.0   & -- \\
    \multicolumn{1}{c|}{} &       & R-C3D~\cite{xu2017r} & ICCV2017 & --    & 54.5  & 51.5  & 44.8  & 35.6  & 28.9  & --    & --    & 43.1  \\
    \multicolumn{1}{c|}{} &       & CBR~\cite{gao2017cascaded} & BMVC2017 & TS    & 60.1  & 56.7  & 50.1  & 41.3  & 31.0  & 19.1  & 9.9   & 47.8  \\
    \multicolumn{1}{c|}{} &       & TAL~\cite{chao2018rethinking} & CVPR2018 & I3D   & 59.8  & 57.1  & 53.2  & 48.5  & 42.8  & \textbf{33.8 } & 20.8  & -- \\
    \midrule
    \multicolumn{2}{c|}{\multirow{5}[1]{*}{Actionness}} & CDC~\cite{shou2017cdc} & CVPR2017 & --    & --    & --    & 40.1  & 29.4  & 23.3  & 13.1  & 7.9   & -- \\
    \multicolumn{2}{c|}{} & SSN~\cite{zhao2017temporal} & ICCV2017 & TS    & 60.3  & 56.2  & 50.6  & 40.8  & 29.1  & --    & --    & 47.4  \\
    \multicolumn{2}{c|}{} & BSN~\cite{lin2018bsn} & ECCV2017 & TS    & --    & --    & 53.5  & 45.0  & 36.9  & 28.4  & 20.0  & -- \\
    \multicolumn{2}{c|}{} & BMN~\cite{lin2019bmn} & ICCV2019 & TS    & --    & --    & 56.0  & 47.4  & 38.8  & 29.7  & 20.5  & -- \\
    \multicolumn{2}{c|}{} & BMN~\cite{lin2019bmn} & ICCV2019 & I3D   & 59.3  & 57.5  & 56.4  & 47.9  & 39.2  & 30.2  & 21.2  & 52.1  \\
    \midrule
    \multicolumn{2}{c|}{Combination} & MGG~\cite{liu2019multi} & CVPR2019 & I3D   & --    & --    & 53.9  & 46.8  & 37.4  & 29.5  & \textbf{21.3 } & -- \\
    \midrule
    \multicolumn{2}{c|}{--} & A2Net & --    & I3D   & 61.1  & 60.2  & \textbf{58.6 } & \textbf{54.1 } & \textbf{45.5 } & 32.5  & 17.2  & \textbf{55.9 } \\
    \bottomrule[2pt]
    \end{tabular}%
  \label{tab:comp-soats-thumos}%
\vspace{-0.3cm}
\end{table*}%

The feature extracted from raw video frames is referred to the spatial feature, which reveals the appearance information of the video. The feature extracted from optical flows is referred to the temporal feature, which is sensitive to motion. The proposed A2Net concatenates the raw spatial features and temporal features to become aware of both appearance cues and motion cues. We train A2Net with only the spatial features or the temporal features to verify the contribution of each modality. A2Net fuses these two kinds of features via concatenation, which is an early fusion manner. In contrast, we can learn two localization networks from spatial features and temporal features individually, obtain two localization results, and finally fuse these two results via average, which is called the late fusion manner. Table \ref{tab_abl_feature} reports the experiment results. With threshold $0.1$:$0.1$:$0.7$, feature fusion methods (A2Net and \emph{LFuse}) exceed both \emph{Spa} and \emph{Tem}, indicating the importance of combining spatial features and temporal features. Among these two fusion manners, the \emph{LFuse} performs inferior to the \emph{early fuse} method, A2Net, by a margin of $2.0$. Thus, A2Net adopts the early fuse pipeline.

\subsection{Comparison with state-of-the-art methods}

We compare A2Net with recent state-of-the-art methods on THUMOS14 and ActivityNet v1.3 datasets. Table \ref{tab:comp-soats-thumos} summarizes the comparison performances on the THUMOS14 dataset. In order to make a thorough comparison, we categorize existing action localization methods into three groups: early researches, anchor-based methods, and actionness-guided methods. As feature representation influences the performance, we report the feature used by each method, including iDT \cite{wang2013action}, PSDF \cite{yuan2016temporal}, TS \cite{wang2015towards} C3D \cite{tran2015learning}, P3D \cite{qiu2017learning} and I3D \cite{carreira2017quo}.

Early researches explore effective solutions in diverse directions, including the gated recurrent unit \cite{buch2017end, buch2017sst}, the reinforcement learning \cite{yeung2016end}, effective optimization \cite{yuan2017temporal, yuan2016temporal}, sub-actions \cite{hou2017real} etc. Although the performances of these early works are not high enough (e.g., the strongest method SS-TAD \cite{buch2017end} achieves mAP@0.5=$29.2$), they provide valuable guidance for subsequent researches.

The anchor-based pipeline is a well-explored research direction, including one-stage methods \cite{lin2017single, long2019gaussian} and two-stage methods \cite{gao2017turn, dai2017temporal, xu2017r, gao2017cascaded, chao2018rethinking}. Among these methods, the one-stage method GTAN \cite{long2019gaussian} shows strong performance under the threshold of $0.1$ and $0.2$. A potential reason is that GTAN adopts more default anchors and is prone to capture more action instances. In precise, GTAN adopts an $8$-level hierarchical network and the temporal length for each level is $\{2^{8}, 2^{7}, ..., 2^{1}\}$. In contrast, A2Net adopts a $6$-level hierarchical network and the temporal length for each level is $\{2^{6}, 2^{5}, ..., 2^{1}\}$. Among these methods, the two-stage method TAL \cite{chao2018rethinking} performs well under the threshold of $0.6$ and $0.7$. One potential reason is that its receptive field alignment module adopts a series of $9$ sub-networks to accurately model the structure for each default anchor, thus generates more precise action localization results. In comparison, A2Net shows superior performance than GTAN \cite{long2019gaussian} under threshold $0.3$:$0.5$, and A2Net exceeds TAL \cite{chao2018rethinking} under threshold $0.1$:$0.5$.

The actionness-guided pipeline is another well-explored research direction, including early researches CDC \cite{shou2017cdc} and SSN \cite{zhao2017temporal},
\begin{table}[htbp]
	\centering
	\small
	\caption{Comparison with some state-of-the-art methods on the THUMOS14 test dataset, measured by average-mAP with threshold $0.3$:$0.1$:$0.7$.}
	\begin{tabular}{c|ccccc}
		\toprule[2pt]
		& CDC   & BSN   & MGG   & BMN   & \multirow{2}[0]{*}{A2Net} \\
		& \cite{shou2017cdc}   & \cite{lin2018bsn}   & \cite{liu2019multi}   & \cite{lin2019bmn}   &  \\
		\midrule[1pt]
		average mAP & 22.8  & 36.8  & 37.8  & 38.5  & \textbf{41.6} \\
		\bottomrule[2pt]
	\end{tabular}%
	\label{tab:cmp-thumos}%
\vspace{-0.3cm}
\end{table}%
as well as recent researches BSN \cite{lin2018bsn} and BMN \cite{lin2019bmn}. Among these methods, BMN \cite{lin2019bmn} shows strong performance. The good performance attributes to the flexibility of the bottom-up actionness-guided pipeline, which is not constrained by the default anchors. Because BMN uses the two-stream feature \cite{wang2015towards} while many recent methods \cite{chao2018rethinking, liu2019multi} use I3D feature \cite{carreira2017quo}, we verify the performance of BMN with I3D feature. Under threshold $0.5$, the I3D feature slightly exceeds the two-stream feature. Although BMN \cite{lin2019bmn} is a good actionness-guided method, A2Net exceeds BMN \cite{lin2019bmn} under threshold $0.3$:$0.6$. This demonstrates that A2Net is flexible enough to tackle actions with different duration.

From the second last row of Table \ref{tab:comp-soats-thumos}, we can find that MGG \cite{liu2019multi} performs superior to A2Net under threshold $0.7$. The main reason is that MGG design a temporal boundary adjustment module in the inference time. In contrast, A2Net follows most previous methods \cite{long2019gaussian, chao2018rethinking}, uses a simple inference strategy and achieves comparable performance.Under threshold 0.3:0.6, A2Net performs superior to MGG. Although MGG \cite{liu2019multi} combines both the sliding-window pipeline and the actionness-guided pipeline, these two modules are learned separately. In contrast, A2Net integrates the anchor-based mechanism and the anchor-free mechanism into a unified framework. Further, A2Net takes advantage of the complementarity between the anchor-free module and the anchor-based module to robustly tackle actions with various duration. These two points make A2Net perform better than MGG \cite{liu2019multi}.

Apart from the state-of-the-art methods discussed above, we notice that there are three strong competitors P-GCN \cite{zeng2019graph}, C-TCN \cite{li2019deep} and TGM \cite{piergiovanni2018temporal}. P-GCN \cite{zeng2019graph} adopts BSN \cite{lin2018bsn} to generate action proposals and achieves mAP@$0.5$=$49.1$. If we replace BSN \cite{lin2018bsn} with TAG \cite{zhao2017temporal} to generate action proposals, the action localization results will drop $8.3$ points using flow feature or drop $6.1$ points using RGB feature. C-TCN \cite{li2019deep} uses data augmentation (including \textit{random move} and \textit{random crop}) to improve the performance from $37.4$ to $52.1$, under the metric mAP@$0.5$. Although TGM \cite{piergiovanni2018temporal} achieves mAP@$0.5$=$53.5$, it finetunes the I3D \cite{carreira2017quo} model on the THUMOS dataset before extracting features and it specifically adjusts parameters for each category\footnote {See \url{https://github.com/piergiaj/tgm-icml19/tree/master/thumos}.}. In contrast, most of the researches (including A2Net) extract features without finetuning the original video recognition model and adopt the same parameters for all categories within a dataset. From these analyses, we can realize that it is unfair to directly compare A2Net with these methods. As the core contribution of this work is the complementarity between the anchor-free module and the anchor-based module, and A2Net shows superior performance than many recent works reported in Table \ref{tab:comp-soats-thumos}. We think the performance of A2Net is promising.

\begin{figure*}[tbp]
	\graphicspath{{figures/}}
	\centering
	\includegraphics[width=0.9\linewidth]{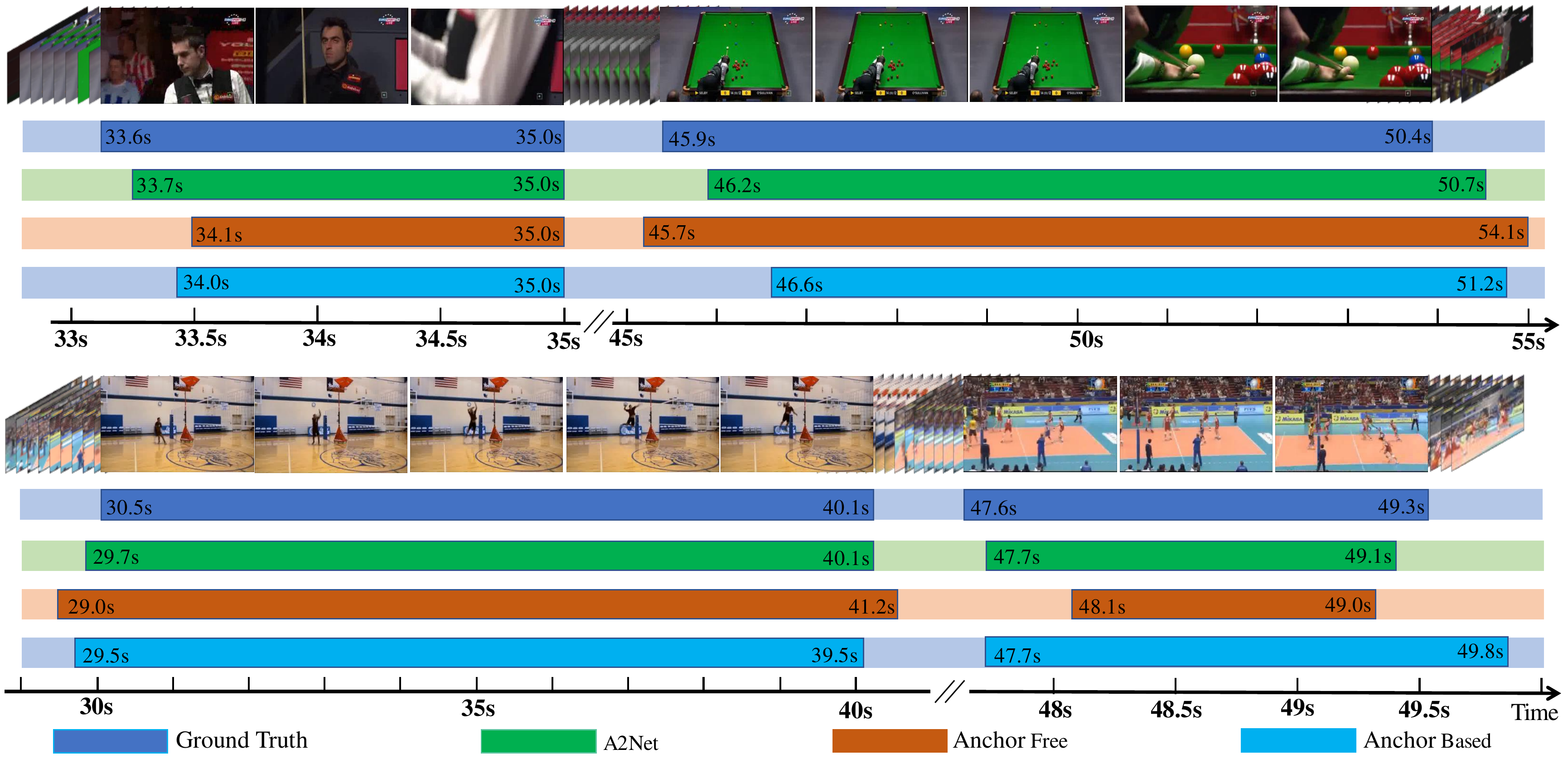}
	\caption{Visualization of two action localization results for A2Net, anchor-free method and anchor-based method on THUMOS14 dataset.}
	\label{fig5_visualization}
\vspace{-0.3cm}
\end{figure*}

In Table \ref{tab:comp-soats-thumos}, although A2Net shows the best performance under the official evaluation metric mAP@$0.5$, someone may argue that A2Net is not always the best one under different thresholds. To make a comprehensive evaluation of the performance, we follow the evaluation metric on ActivityNet dataset \cite{caba2015activitynet} and report the average mAP. In Table \ref{tab:comp-soats-thumos}, most methods report performances under threshold $0.1$:$0.5$, thus we calculate average mAP with respect to the threshold $0.1$:$0.1$:$0.5$. A2Net shows the best average mAP in Table \ref{tab:comp-soats-thumos}. Moreover, some methods report mAP under threshold $0.3$:$0.7$. We report the average mAP respect to the threshold $0.3$:$0.1$:$0.7$ in Table \ref{tab:cmp-thumos}, and A2Net shows the best performance again.

From the above discussions, A2Net shows obvious superiority over existing action localization methods. The reason for such superiority mainly lies in two aspects. Firstly, the inherent complementarity of the anchor-based module and the
\begin{table}[tbp]
  \caption{Comparison with state-of-the-art methods on the ActivityNet v1.3 validation dataset, measured by mAP.}
  \label{tab_comp_anet}
  \centering
  \small
  \begin{tabular}{c|cccc}
\toprule[2pt]
Approach                    	&0.50	&0.75	&0.95	&avg-mAP	\\
\midrule[1pt]
R-C3D~\cite{xu2017r}            &26.80	&--  	&-- 	&12.70	\\
TAL~\cite{chao2018rethinking} 	&38.23	&18.30	&1.30	&20.22	\\
CDC~\cite{shou2017cdc}  		&45.30	&26.00	&0.20	&23.80	\\
Dai et al.~\cite{dai2017temporal}&36.44	&21.15	&3.90	&--	    \\
TAG~\cite{xiong2017pursuit}     &39.12	&23.48	&5.49	&23.98	\\
P-GCN~\cite{zeng2019graph}      &42.90	&28.14	&2.47	&26.99	\\
BSN~\cite{lin2018bsn}   		&52.50	&33.53	&8.85	&33.72	\\
BMN~\cite{lin2019bmn}   		&50.07	&\textbf{34.78}	&8.29	&33.85	\\
GTAN~\cite{long2019gaussian}	&\textbf{52.62}	&34.14	&\textbf{8.91}	&\textbf{34.31}	\\
\midrule[1pt]
A2Net 	                        &43.55	&28.69	&3.70	&27.75	\\
\bottomrule[2pt]
  \end{tabular}
\vspace{-0.3cm}
\end{table}
proposed anchor-free module determines that the prediction results from these two modules can cooperate with each other and precisely locate out action instances. Secondly, A2Net is a complete framework based on a shared \emph{Hierarchical Feature Module}. It can simultaneously learn anchor-free and anchor-based modules in an end-to-end trainable manner. Because of the cooperation of anchor-based module and anchor-free module, A2Net can not only dispose of action instances with medium length but also tackle the large variation of action instances, providing a simple but effective framework for the action localization task.

Table \ref{tab_comp_anet} reports the performance of A2Net on ActivityNet v1.3. It can be found that A2Net performs better than anchor-based methods R-C3D \cite{xu2017r}, TAL \cite{chao2018rethinking} and actionness-guided methods CDC~\cite{shou2017cdc}, TAG~\cite{xiong2017pursuit}. We can also find that A2Net performs inferior to BSN~\cite{lin2018bsn}, BMN~\cite{lin2019bmn} and GTAN~\cite{long2019gaussian}. Actually, both BSN~\cite{lin2018bsn} and BMN~\cite{lin2019bmn} are intended for generating action proposals. They adopt the detection-by-classification pipeline to perform action localization. The classification labels are obtained from a well-performed video recognition method \cite{zhao2017cuhk} on the ActivityNet dataset, which is a heavy and complicated model for video recognition. Among these methods, GTAN~\cite{long2019gaussian} achieves the best performance. However, GTAN~\cite{long2019gaussian} requires more channels and layers than A2Net. The large computation turns GTAN into a bit complicated for practical applications. Given its simplicity, A2Net shows comparable performance on ActivityNet v1.3, making A2Net appropriate to serve as a foundation for further high-level video processing researches.
% \blue{In detail, the computation cost for GTAN and A2Net is $5.04$G FLOPS and $2.48$G FLOPs, respectively.}

In Figure \ref{fig5_visualization}, we visualize two action localization results on videos from the THUMOS14 dataset. From the four action instances from these two videos,
\begin{table}[tbp]
  \centering
  \caption{Comparison of model efficiency, in terms of inference time and FLOPs. We report mAP under threshold $0.5$ on THUMOS14 dataset. "*" indicates the weakly supervised method.}
  \setlength{\tabcolsep}{2pt}
    \begin{tabular}{c|cccc|c}
    \toprule[2pt]
          & A2Net & \multicolumn{1}{l}{SSAD \cite{lin2017single}} & \multicolumn{1}{l}{BMN \cite{lin2019bmn}} & \multicolumn{1}{l|}{GTAN \cite{long2019gaussian}} & \multicolumn{1}{l}{3C-Net* \cite{narayan20193c}} \\
    \midrule
    Infe. Time / ms & 13.24 & 2.37  & 16.96 & 15.14 & 2.50 \\
    FLOPs / G & 1.24  & 0.45  & 91.25 & 2.51  & 0.40 \\
    \midrule
    mAP   & 45.5  & 24.6  & 38.8  & 38.8  & 26.6 \\
    \bottomrule[2pt]
    \end{tabular}%
  \label{tab-model-efficiency}%
  \vspace{-0.3cm}
\end{table}%
it can be found that neither the anchor-free method nor the anchor-based method can achieve accurate action localization performance individually. However, A2Net is able to precisely locate these action instances. Essentially, the anchor-based method can only generate coarse action localization boundaries for short or long action instances, while the accurate localization for medium length action instances is challenging for the anchor-free module. It is the complementarity between these two modules that lead to the good performance of A2Net.

\subsection{Model Efficiency}
\label{subsection-efficiency}

Model efficiency is an important factor for practical applications. We compare the model efficiency among the proposed A2Net and state-of-the-art supervised action localization methods \cite{lin2017single, long2019gaussian, lin2019bmn}. Besides, some weakly supervised action localization methods only rely on video-level classification labels \cite{narayan20193c} or single-frame supervision \cite{ma2020sf}. They are efficient for annotating and fast for inferencing, which are also compared here. The results are shown in Table \ref{tab-model-efficiency}. Compared with supervised methods, A2Net requires medium computation FLOPs and costs medium inference time, but achieves superior localization performance. BMN \cite{lin2019bmn} requires less computations than A2Net, but its inference time is longer. The reason is that BMN uses the boundary matching layer to construct features and calculate confidence scores for all proposals,
\begin{table}[tbp]
  \caption{Performance of anchor-free (AF) module to assist proposal generation, measured by mAP. R-C3D*~\cite{swfaster2rc3d} is a re-production of original R-C3D.}
  \label{tab_comp_thumos14}
  \centering
  \footnotesize
  %\small
  \begin{tabular}{c|ccccccc}
 \toprule[2pt]
tIoU	                        &0.1	&0.2	&0.3	&0.4	&0.5	&0.6	&0.7\\
\midrule[1pt]
R-C3D~\cite{xu2017r}            &54.5	&51.5	&44.8	&35.6	&28.9	&--		&--		\\
R-C3D*~\cite{swfaster2rc3d}     &50.6 	&49.6 	&46.9 	&40.8 	&32.5	&24.6	&14.3   \\
R-C3D* + AF                     &\textbf{57.8} 	&\textbf{57.1} 	&\textbf{53.4} 	&\textbf{47.4} 	&\textbf{37.0}	&\textbf{28.7}	&\textbf{17.3}	\\
\bottomrule[2pt]
  \hline
  \end{tabular}
\vspace{-0.1cm}
\end{table}
\begin{table}[tbp]
  \centering
  \caption{Performance of anchor-free (AF) module to assist proposal generation, measured by AR@AN and AUC. R-C3D*~\cite{swfaster2rc3d} is a re-production of original R-C3D.}
    \begin{tabular}{c|ccccc|c}
    \toprule[2pt]
          & \multicolumn{5}{c|}{AR@AN} & \multirow{2}[2]{*}{AUC} \\
          & @50   & @100  & @200  & @500  & @1000 &  \\
    \midrule[1pt]
    R-C3D & 28.09 & 40.06 & 52.34 & 67.35 & 75.18 & 61.72 \\
    R-C3D + AF & 38.55 & 49.84 & 59.66 & 72.11 & 79.01 & 67.19 \\
    \bottomrule[2pt]
    \end{tabular}%
  \label{tab_af_proposal}%
  \vspace{-0.3cm}
\end{table}%
which is time-consuming. Although the weakly-supervised method 3C-Net \cite{narayan20193c} is efficient in computation, there is $18.9$ performance gap between 3C-Net and the proposed A2Net, for mAP under threshold $0.5$. In summary, the proposed A2Net is a proper choice when simultaneously considering efficiency and effectiveness.

\subsection{Anchor-free Module for Generating Action Proposals}
\label{subsection-af-proposal}
The above experiments have demonstrated the effectiveness of integrating the proposed anchor-free module with a traditional one-stage anchor-based module for action localization. As discussed in \ref{af_proposal}, we carry on experiments to verify the performance of the anchor-free module for generating proposals. We start from R-C3D~\cite{xu2017r}, an end-to-end trainable two-stage action localization method. We found a re-production of R-C3D \cite{swfaster2rc3d}, which performs better. Based on this re-produced code \cite{swfaster2rc3d}, we integrate the anchor-free module into the \textit{Action Proposal Network} and train the complete network in an end-to-end manner.

Table \ref{tab_comp_thumos14} reports the action localization performance. It can be noted that, under threshold $0.1$:$0.7$, the anchor-free mechanism consistently brings improvement (from $3$ points to $7$ points) over a well-performed implementation~\cite{swfaster2rc3d}. Table \ref{tab_af_proposal} reports the proposal generation performance. Under metrics AR@AN and AUC, the proposed anchor-free module consistently improves the quality of proposals. The improvement comes from the complementarity between the anchor-free module and the anchor-based module. The anchor-based module learns the offset with respect to the default anchors, making it is sensitive to relative distances. Meanwhile, the anchor-free module directly learns the distances to action boundaries, making it is flexible. When we incorporate the anchor-free module into the conventional two-stage action localization methods, the learned features are more expressive and the action proposal network improves a lot.

\section{Conclusion}
In this paper, we introduce an anchor-free mechanism for temporal action localization task, aiming to deal with the diverse duration of action instances. The anchor-free mechanism represents an action instance via a temporal point and its distances to the starting and ending boundaries, avoids the prior bias about the localization and duration. Then, we incorporate the proposed anchor-free module with a traditional anchor-based method and present a unified framework, namely A2Net. Comprehensive experiments demonstrate the complementarity between the anchor-based module and the proposed anchor-free module. Based on this complementarity, A2Net is capable of precisely localize action instances with various duration. It achieves promising performance on THUMOS14 dataset, providing a simple and effective baseline for subsequent researches.

Although A2Net only shows competitive performance on ActivityNet v1.3, the complementarity between the anchor-free module and the anchor-based module is still verified by Table \ref{tab_complenmentarity_anet} and Table \ref{tab_comp_anet}. Furthermore, we demonstrate that the anchor-free module can hep the \textit{Action Proposal Network} to achieve better performance in \ref{af_proposal} and \ref{subsection-af-proposal}, which is inspiring for subsequent action localization researches.

Currently, the majority of the action localization methods (including A2Net) first extract features from the video sequences and then directly learn from these features. Learning feature extraction and action localization in an end-to-end framework is a worth-trying way for future research. Besides, the proposed A2Net demonstrates that the anchor-free module is an important component of action localization methods. This discovery can bring inspiration to related researches, e.g. video re-localization~\cite{feng2018video} and video moment localization~\cite{anne2017localizing}.

%\bibliographystyle{IEEEtran}
%\bibliography{egbib}


%\vspace{-1.5cm}
\begin{IEEEbiography}[{\includegraphics[width=1in,height=1.25in,clip,keepaspectratio]{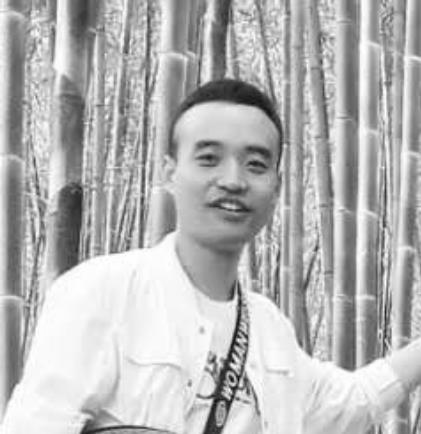}}]{Le Yang} received his B.E. degree from Northwestern Polytechnical University, Xi'an, China, in 2016. He is currently a Ph.D. candidate in the School of Automation at Northwestern Polytechnical University. His research interests include video action localization and video object segmentation.
\end{IEEEbiography}

%\vspace{-3.5cm}
\begin{IEEEbiography}[{\includegraphics[width=1in,height=1.25in,clip,keepaspectratio]{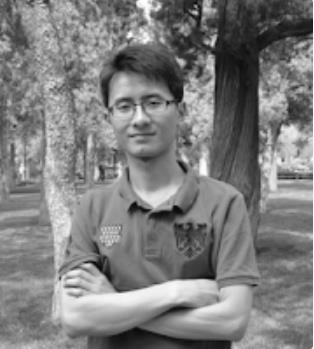}}]{Houwen Peng}
is a researcher working on computer vision and deep learning at Microsoft Research as of 2018. Before that he was a senior engineer at Qualcomm AI Research. He received Ph.D. from NLPR, Instituation of Automation, Chinese Academy of Sciences in 2016. From 2015 to 2016, he worked as a visiting research scholar at Temple University. His research interest includes neural architecture design and search, video object tracking, segmentation and detection, video moment localization, saliency detection, etc.
\end{IEEEbiography}

%\vspace{-3.5cm}
\begin{IEEEbiography}[{\includegraphics[width=1in,height=1.25in,clip,keepaspectratio]{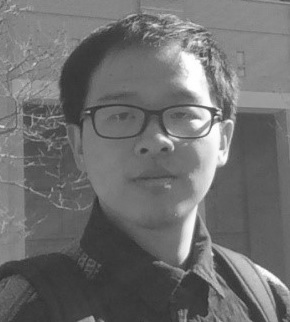}}]{Dingwen Zhang} received his Ph.D. degree from the Northwestern Polytechnical University, Xi'an, China, in 2018. He is currently an associate professor in the School of Mechano-Electronic  Engineering, Xidian University. From 2015 to 2017, he was a visiting scholar at the Robotic Institute, Carnegie Mellon University. His research interests include computer vision and multimedia processing, especially on saliency detection, video object segmentation, and weakly supervised learning.
\end{IEEEbiography}

%\vspace{-3.5cm}
\begin{IEEEbiography}[{\includegraphics[width=1in,height=1.25in,clip,keepaspectratio]{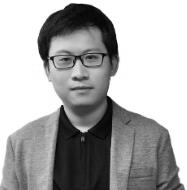}}]{Jianlong Fu}
(M18) received the Ph.D. degree in pattern recognition and intelligent system from the Institute of Automation, Chinese Academy of Science, in 2015. He is currently a Lead Researcher with the Multimedia Search and Mining Group, Microsoft Research Asia (MSRA). He has authored or coauthored more than 40 papers in journals and conferences, and one book chapter. His current research interests include computer vision, computational photography, vision, and language. He received the Best Paper Award from ACM Multimedia 2018, and has shipped core technologies to a number of Microsoft products, including Windows, Office, Bing Multimedia Search, Azure Media Service, and XiaoIce. He is an Area Chair of ACM Multimedia 2018, ICME 2019. He serves as a Lead organizer and a Guest Editor for the IEEE Transactions on Pattern Analysis and Machine Intelligence Special Issue on Fine-grained Categorization.
\end{IEEEbiography}

%\vspace{-3.5cm}
\begin{IEEEbiography}[{\includegraphics[width=1in,height=1.25in,clip,keepaspectratio]{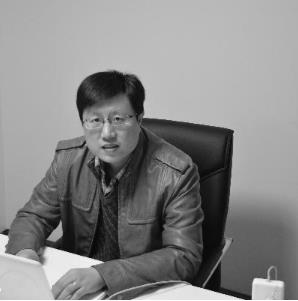}}]{Junwei Han}
is currently a Professor in the School of Automation, Northwestern Polytechnical University. His research interests include computer vision, pattern recognition, remote sensing image analysis, and brain imaging analysis. He has published more than 70 papers in top journals such as IEEE TPAMI, TNNLS, IJCV, and more than 30 papers in top conferences such as CVPR, ICCV, MICCAI, and IJCAI. He is an Associate Editor for several journals such as IEEE TNNLS and IEEE TMM.
\end{IEEEbiography}

\end{document}